\documentclass[11pt, oneside]{article}    

\usepackage{booktabs}
\usepackage{multirow}
\usepackage{dsfont}
\usepackage[table,xcdraw]{xcolor}
\usepackage{geometry}
\usepackage{graphicx}
\usepackage{color}
\usepackage{xcolor}
\usepackage{amssymb}
\usepackage{hyperref}
\usepackage{adjustbox}
\hypersetup{
    colorlinks=true,
    linkcolor=black,
    citecolor=black,
    urlcolor=blue,
}

\usepackage{natbib}
\setcitestyle{authoryear,open={(},close={)}}

\title{Bridging Sim2Real Gap Using Image Gradients for the Task of End-to-End Autonomous Driving}
\author{
  \small \textbf{\scriptsize Unnikrishnan R Nair}\\
  \small \texttt{\scriptsize unnikrishnan.r@olaelectric.com,}
   \small \texttt{\scriptsize Ola Electric}
  \and
  \small \textbf{\scriptsize Sarthak Sharma}\\
  \small \texttt{\scriptsize sarthak.sharma1@olaelectric.com,}
   \small \texttt{\scriptsize Ola Electric}
  \and
  \small \textbf{\scriptsize Udit Singh Parihar}\\
  \small \texttt{\scriptsize udit.parihar@olaelectric.com,}
   \small \texttt{\scriptsize Ola Electric}
  \and
  \small \textbf{\scriptsize Midhun S Menon}\\
  \small \texttt{\scriptsize midhun.s@olaelectric.com,}
   \small \texttt{\scriptsize Ola Electric}
  \and
  \small \textbf{\scriptsize Srikanth Vidapanakal}\\
  \small \texttt{\scriptsize srikanth.vidapanakal@olaelectric.com,}
   \small \texttt{\scriptsize Ola Electric}
}

\date{\today}

\begin{document}
\maketitle




\subsection*{Keywords}
Reinforcement Learning, Sim2Real Transfer, Racecar, Computer Vision

\section{Introduction}

While simulations are vital for developing and testing robotic agents, the ability to transfer robotics skills learned in simulation to reality is equally challenging. Simulation to real-world transfer, popularly known as \textit{Sim2Real} transfer, is an indispensable line of research for utilizing knowledge learned from simulated data to derive meaningful inferences from real-world observations and function in actual operational environments. \textit{Sim2Real} is critical for developing Autonomous Vehicles (AV) and other field-deployed intelligent robotic systems.

\begin{figure*}[t!]
  \includegraphics[width=\textwidth,height=6cm]{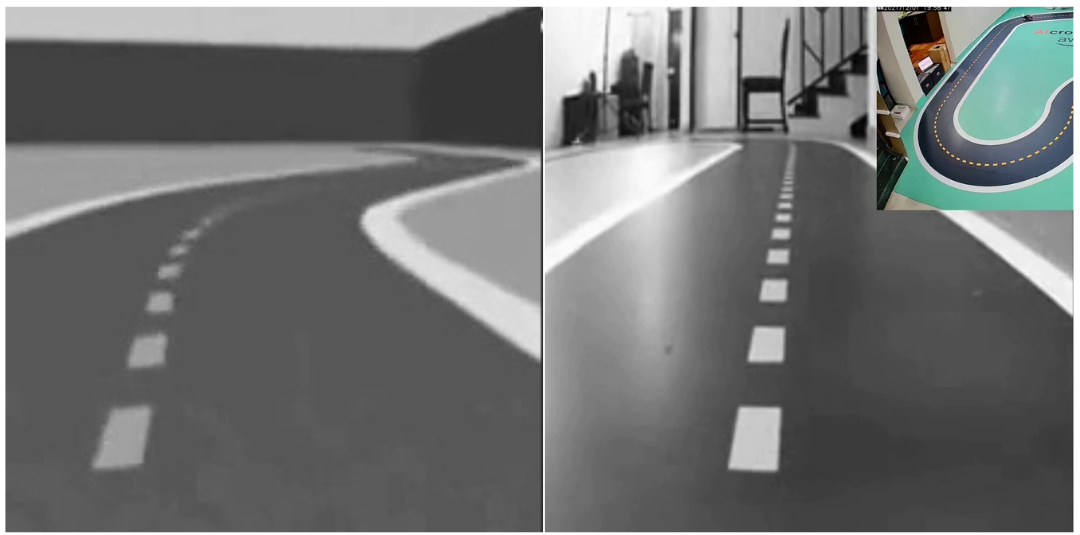}
  
  \caption{ \textbf{Simulator and Real World Environment}: \textit{Left} We show the image captured from the greyscale camera mounted on the vehicle of the simulator. \textit{Right} We show the image captured by the grey scale camera mounted on top of the vehicle of the real track. \textit{Top-right} We show the image of the actual track.}
  \label{fig:tracks}
\end{figure*}

The NeurIPS 2021 AWS DeepRacer challenge is a part of a series of competitions in the area of AV called The AI Driving Olympics (AI-DO). In this competition, the task is to train a Reinforcement Learning (RL) agent (i.e., an autonomous car) that learns to drive by interacting with a virtual environment (a simulated track). The agent achieves this by taking action based on previous observations and a current state to maximize the expected reward. We test the trained agent's driving performance when maneuvering an AWS DeepRacer car on a real-world mock track. The driving performance metric is time to lap completion without going off the same in the context of AWS DeepRacer competition. In Fig~\ref{fig:tracks}, we show the images of the simulated (left) and actual (right) track as captured by the camera mounted on the vehicle with matching extrinsic parameters.
As the car navigates through the environment (in a simulator or in real world), it is allowed to observe a greyscale monocular image from a front-facing camera mounted on the vehicle. At any instance, the car can take a set of actions from a discrete action space $A=\{left-high, left-med,  front, right-med, right-high\}$, where $med$ and $high$ capture the intensity of taking a particular direction. The vehicle's speed remains constant and is not part of the action space.

Given the limited access to the vehicle's controls, we modeled our approach as a classification task over the action space $A$ in an imitation learning setup, which takes the current image observation at any given time $t$. While driving in the simulator environment \textit{S}, we record the action $a_{s,t} \in A$ taken at every time instant $t$, along with the corresponding camera image $I_{s,t}$. Similarly, while driving in the real environment \textit{R}, we record the corresponding camera image $I_{r,t}$. The task is to obtain a probability distribution over the set of actions $A$, $\mathds{P}(y=a_{r,t} | I_{r,t})$, with the simulator recorded data $D_{s} = \{ (a_{s,t},I_{s,t})\}$ as the training set.

\subsection{Related Work}

\textbf{Imitation Learning}: Imitation learning (IL) has seen recent surge for the task of autonomous driving (\cite{zhang2016query,pan2017agile}). IL uses expert demonstrations to directly learn a policy that maps states to actions. The pionerring work of (\cite{pomerleau1988alvinn}) intoduced IL for self-driving  - where a direct mapping from the sensor data to steering angle and acceleration is learned. (\cite{zeng2019end,viswanath2018end}) also follow the similar approach of going from sensor data to throttle and steering. 
With the advent of high end driving simulators like (\cite{dosovitskiy2017carla}), approaches like (\cite{codevilla2018end}) exploit conditional models with additional exploit conditional models with additional high-level commands such as \textit{continue, turn-left, turn-right}. Methods like (\cite{muller2018driving}) use intermediate representations derived from sensor data, which is then converted to steering commands. 
\\

\noindent \textbf{Sim2Real transfer}: Simulation to real-world transfer, popularly known as \textit{Sim2Real} transfer has contributed immensely in training of models for different robotics tasks. \textit{Sim2Real} transfer provides several benefits like bootstrapping, hardware in loop optimization and aiding when there is a real-world data starvation. However, \textit{Sim2Real} methods suffer from domain gap between the simulated and real world. Approaches have explored domain randomization (\cite{amiranashvili2021pre}), explicit transferable abstraction (\cite{julian2020scaling}) and domain adaption vis GANs (\cite{zhu2017unpaired}) to bridge the \textit{Sim2Real} gap.



\section{Methods}

\subsection{Algorithm}
We model our approach as a classification task over the action space $A$ in an imitation learning setup. Specifically, while driving in the simulator environment \textit{S}, we record the action $a_{s,t} \in A$ taken at every time instant $t$, along with the corresponding camera image $I_{s,t} \in \mathbb{R}^{H \times W }$. Motivated by (\cite{tolstikhin2021mlp}), we train a MLP-Mixer $f_{\theta}$, where $\theta$ are the parameters of the network. The model takes in as input $I_{g,s,t}$ where $I_{g,s,t} = g(I_{s,t})$, where $g$ is the gradient computation function. The output of the model $f_{\theta}$ is scored actions :

\[ 
\phi (\mathbf{s_{a}}|I_{s,t}) = \frac{exp(f_\theta(I_{g,s,t}))}{\sum_{i=1}^{5} exp(f_{\theta}^i(g(I_{s,t}))}
\]

The loss term for training this stage is the cross entropy
between the predicted scores and ground truth scores:

\[
L = L_{CE}(\phi (\mathbf{s_{a}}|I_{s,t}),  \mathbf{a}_{s,t})
\]

At test time, we receive the image $I_{r,t}$ recorded from the DeepRacer car moving in the real world. Gradient computed image $I_{g,r,t}$ is passed through the above trained network $f_{\theta}$, to obtain the scores on the action space $\mathbf{s_{a}}$. The most likely action is then chosen and relayed to the vehicle to execute.

\begin{figure*}[t!]
  \includegraphics[width=\textwidth,height=8cm]{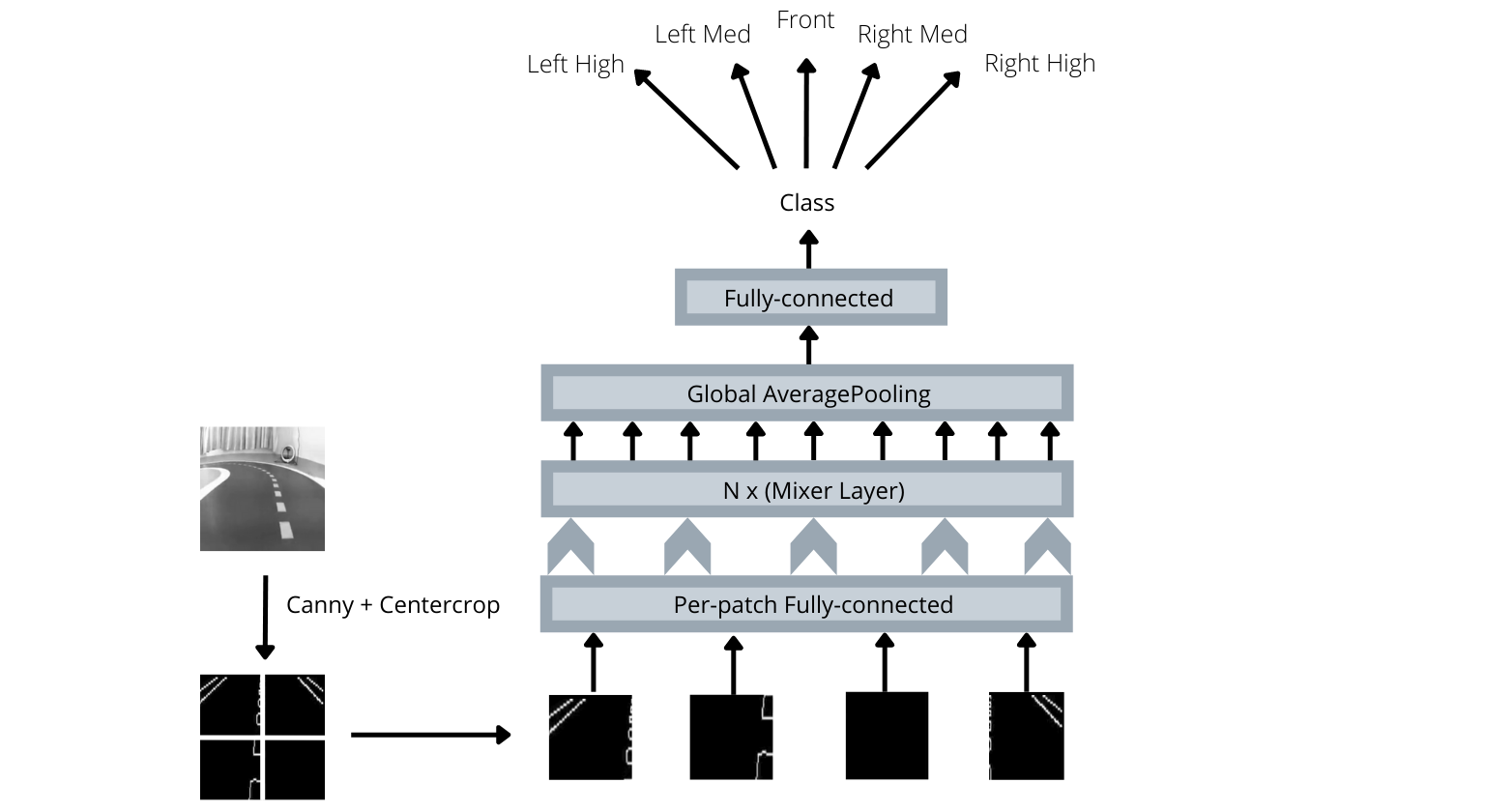}
  
  \caption{ \textbf{Architecture}: We show the architecture diagram of our proposed method. The input $image I_{s,t}$ is passed through a Canny edge detection method to obtain the input $I_{g,s,t}$ which is then fed to the MLP mixer model $f_\theta$ that computes the scores $\mathbf{s}_a$ over the actions $a \in A$. At test time, we get the input image from the real world, $I_{r,t}$, on which image gradients are calculated using Canny edge detection to obtain $I_{g,r,t}$. This is then forwarded through the learned weights $f_\theta$ to obtain the action scores $\mathbf{s}_a$. The most likely action is then chosen and relayed to the vehicle to execute. } 
  \label{fig:arch}
\end{figure*}

We showcase our network architecture in Fig. \ref{fig:arch}. For our experiments, we keep $H=64$, $W=64$ and train using SGD as the optimizer with learning rate of $1e-4$.  Other hyper-parameters include the parameters for the MLP-Mixer (\cite{tolstikhin2021mlp}) which are patch size $= 8$, dimension $= 128$, depth $= 6$.

\subsection{\textit{Sim2Real} Transfer}

\begin{figure*}[t!]
  \includegraphics[width=\textwidth,height=4cm]{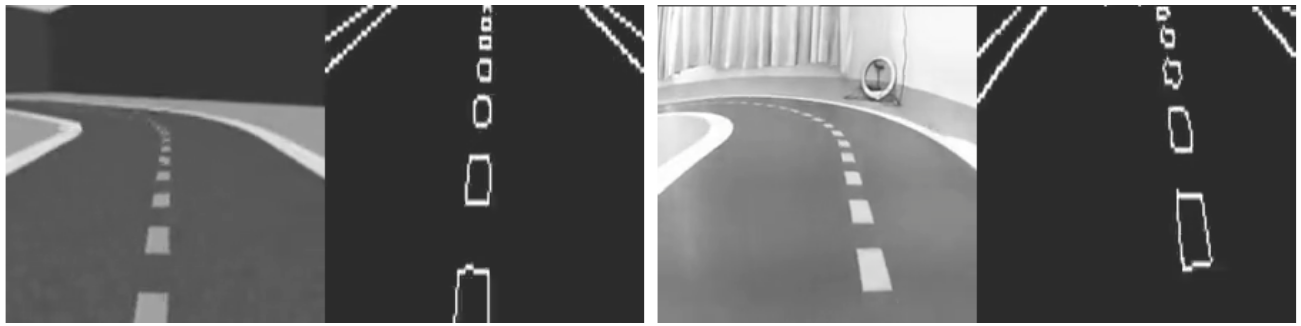}
  
  \caption{ \textbf{Image gradients on simulator and real world images}: \textit{Left} We show the Canny edge detection for simulator image $I_{s,t}$ \textit{Right} We show the Canny edge detection for real world image $I_{r,t}$. For both simulator and real world, we remove the top $20\%$ of the image before computing the image gradients.}
  \label{fig:canny}
\end{figure*}

We achieve \textit{Sim2Real} transfer for our method by passing the input images from simulator and real world through Canny edge detection. Specifically, when we get an input image from any of the real or simulator environment, we pass it through a Canny edge detector to obtain the corresponding image gradients $I_{g,t}$. The training set then updates to $D_{s} = \{ (a_{s,t},I_{g,s,t})\}$. We showcase the results for both the real image and simulated image once after performing Canny edge detection in Fig.\ref{fig:canny}. Image gradient of $100$ along X and $256$ along Y axis was set while running Canny edge detection. We remove unnecessary image regions, which contain irrelevant information for driving, by cropping top $20\%$ of the image before computing the image gradients. 

\section{Results}
To increase robustness, we perform filtering of training data: removing sub-optimal observation and action pairs $(I_{s,t}, a_{s,t})$. Due to this, we were able to achieve robust performance in track completion even when $50\%$ of the commands were randomly changed. The overall runtime of the model was only $2-3$ ms on a modern CPU.

Our method ranked $1st$ on the NeurIPS 2021 AWS DeepRacer challenge, outperforming over $250$ participants from $30+$ teams worldwide.

\section{Discussion}
Automatic dataset filtering has been performed to ensure that the training data only contains the optimal observation, action and reward triplet. For this we have utilized the reward signal provided by the simulator and ensured that the subsequent reward only improves from the previous value. This aids in removing the error that creeps in due to human-expert based data collection.
We have iteratively arrived at the optimal network architecture and the corresponding input image preprocessing. While Canny edge detecion enables \textit{Sim2Real} transfer, it also help in overcoming issues like reflection on the marble floor which was causing the network to select the wrong action. To overcome the low-end spec hardware in the DeepRacer car, we used MLP-Mixer architecture (\cite{tolstikhin2021mlp}) which gave us a runtime of $2-3$ ms.




\subsection{Failed ideas}
Advantage Actor Critic by \cite{a2c} and DQN by \cite{dqn} methods were experimented with, but were discarded due to low score on the simulator track.


\bibliography{citations_olav}
\end{document}